\documentclass{article}
\usepackage{spconf,amsmath,amssymb,epsfig}
\usepackage{hyperref}
\usepackage{xcolor}
\usepackage{booktabs}
\usepackage{multirow}
\usepackage{enumitem}

\def\ChangeFormer{{\textit{ChangeFormer}}}

\title{A Transformer-Based Siamese Network for Change Detection}

\name{Wele Gedara Chaminda Bandara, Vishal M. Patel } 
\address{Johns Hopkins University, Baltimore, Maryland, USA.\\ $\tt \{wbandar1, vpatel36\}@jhu.edu$}

\begin{document}

\maketitle

\begin{abstract}
This paper presents a transformer-based Siamese network architecture (abbreviated by \ChangeFormer) for Change Detection (CD) from a pair of co-registered remote sensing images. Different from recent CD frameworks, which are based on fully convolutional networks (ConvNets), the proposed method unifies hierarchically structured transformer encoder with Multi-Layer Perception (MLP) decoder in a Siamese network architecture to efficiently render multi-scale long-range details required for accurate CD. Experiments on two CD datasets show that the proposed end-to-end trainable \ChangeFormer{} architecture achieves better CD performance than previous counterparts. Our code and pre-trained models are available at \href{https://github.com/wgcban/ChangeFormer}{\color{blue}{$\tt github.com/wgcban/ChangeFormer$}}.
\end{abstract}


\section{Introduction}
Change Detection (CD) aims to detect relevant changes from a pair of co-registered images acquired at distinct times~\cite{bandara2022revisiting}. The definition of \textit{change} may usually vary depending on the application. The changes in man-made facilities (e.g., buildings, vehicles, etc.), vegetation changes, and environmental changes (e.g., polar ice cap melting, deforestation, damages caused by disasters) are usually regarded as {relevant changes}. A better CD model is the one that can recognize these relevant changes while avoiding complex \textit{ irrelevant changes} caused by seasonal variations, building shadows, atmospheric variations, and changes in illumination conditions.

The existing state-of-the-art (SOTA) CD methods are mainly based on deep convolutional networks (ConvNets) due to their ability to extract powerful discriminative features. Since it is essential to capture \textit{long-range contextual information} within the spatial and temporal scope to identify relevant changes in multi-temporal images, the latest CD studies have been focused on increasing the \textit{receptive field} of the CD model. As a result, CD models with stacked convolution layers, dilated convolutions, and attention mechanisms~\cite{bandara2021spin} (channel and spatial attention) have been proposed \cite{cd_attention}. Even though the attention-based methods are effective in capturing global details, they struggle to relate long-range details in space-time because they use attention to re-weight the bi-temporal features obtained through ConvNets in the channel and spatial dimension.

The recent success of \textit{ Transformers} (i.e., non-local self-attention) in Natural Language Processing (NLP) has led researchers in  applying transformers in various computer vision tasks. Following the transformer design in NLP, different architectures have been proposed for various computer vision tasks, including image classification and image segmentation such as Vision Transformer (ViT), SEgmentation TRansformer (SETR), Vision Transformer using Shifted Windows (Swin), Twins \cite{transformers_review} and SegFormer \cite{segformer}. These Transformer networks have comparatively larger \textit{ effective receptive field (ERF)} than deep ConvNets - providing much stronger context modeling ability between any pair of pixels in images than ConvNets.

Although Transformer networks have a larger receptive field and stronger context shaping ability, very few works have been done on transformers for CD. In a more recent work \cite{transformer_cd}, a transformer architecture is applied in conjunction with a ConvNet encoder (ResNet18) to enhance the feature representation while keeping the overall ConvNet-based feature extraction process in place. \textit{In this paper, we show that this dependency on ConvNets is not necessary, and a hierarchical transformer encoder with a lightweight MLP decoder can work very well for CD tasks.}

\section{Method}
\label{sec:method}
\begin{figure*}[tbh]
    \centering
    \includegraphics[width=\linewidth]{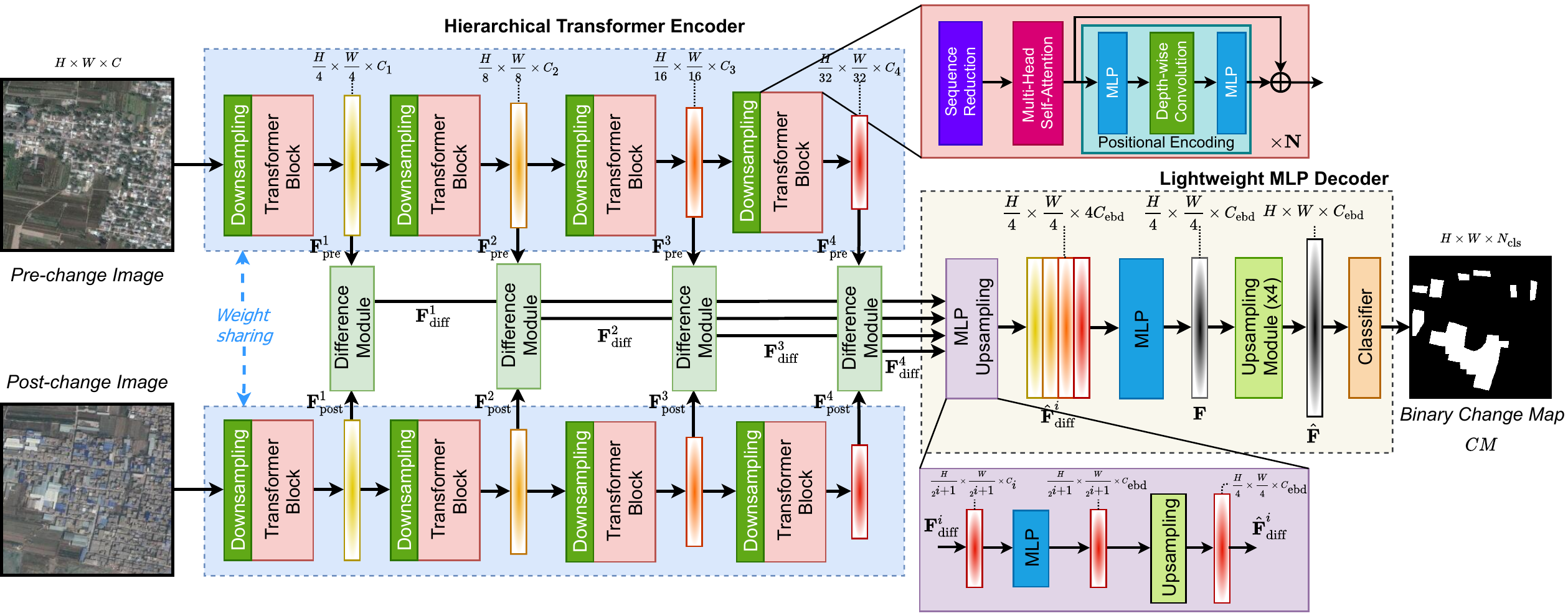}
    \caption{The proposed \ChangeFormer{} network for CD.}
    \label{fig:ChangeFormer}
\end{figure*}

The proposed \ChangeFormer{} network consists of three main modules as shown in Fig. \ref{fig:ChangeFormer}: a hierarchical transformer encoder in a Siamese network to extract coarse and fine features of bi-temporal image, four feature difference modules to compute feature differences at multiple scales, and a lightweight MLP decoder to fuse these multi-level feature differences and predict the CD mask.

\subsection{Hierarchical Transformer Encoder}
Given an input bi-temporal image, the hierarchical transformer encoder generates ConvNet-like multi-level features with high-resolution coarse features and low-resolution fine-grained features required for the CD. Concretely, given a pre-change or post-change images of resolution $H \times W \times 3$, the transformer encoder outputs feature maps $\mathbf{F}_i$ with a resolution $\frac{H}{2^{i+1}} \times \frac{W}{2^{i+1}} \times C_i$, where $i=\{1,2,3,4\}$ and $C_{i+1} > C_i$ which will be further processed through the difference modules followed by MLP decoder to obtain the change map.

\subsubsection{Transformer Block} 
The main building block of the transformer encoder is \textit{self-attention} module. In the original work \cite{original_attention}, self-attention is estimated as:
\begin{equation}
    \text{Attention}(\mathbf{Q},\mathbf{K},\boldsymbol{V}) = \texttt{Softmax} \left( \frac{\mathbf{Q} \mathbf{K}^T}{\sqrt{d_{\text{head}}}}\right) \mathbf{V},
    \label{eq:atten}
\end{equation}
where $\mathbf{Q}$, $\mathbf{K}$, and $\mathbf{V}$ denote Query, Key, and Value, respectively, and have the same dimensions of $HW \times C$. However, the computational complexity of eqn. (\ref{eq:atten}) is $O((HW)^2)$ which prohibits its application on high-resolution images. To reduce the computational complexity of eqn. (\ref{eq:atten}), we adopt the \textit{Sequence Reduction} process introduced in \cite{eff_atten} which utilizes reduction ratio $R$ to reduce the length of the sequence $HW$ as follows:
\begin{equation}
    \hat{\mathbf{S}} = \texttt{Reshape} \left( \frac{HW}{R}, C \cdot R \right) \mathbf{S},
\end{equation}
\begin{equation}
    \mathbf{S} = \texttt{Linear} \left( C \cdot R, C \right) \hat{\mathbf{S}},
\end{equation}
where $\mathbf{S}$ denotes the sequence to be reduced i.e., $\mathbf{Q}$, $\mathbf{K}$, and $\mathbf{V}$,  $\texttt{Reshape}(h, w)$ denotes tensor reshaping operation to the one with
shape of $(h,w)$,  and $\texttt{Linear}(C_{\text{in}}, C_{\text{out}})$ denotes a linear-layer with $C_{\text{in}}$ input channels and $C_{\text{out}}$ output channels. This results in a new set of $\mathbf{Q, K},$ and $\mathbf{V}$ of size $\left( \frac{HW}{R},C\right)$, hence reduces the computational complexity of eqn. (\ref{eq:atten}) to $O((HW)^2/R)$.

To provide \textit{positional information} for transformers, we utilize two MLP layers along with a $3 \times 3$ depth-wise convolutions as follows:
\begin{equation}
    \mathbf{F}_{\text{out}} = \texttt{MLP}(\texttt{GELU}(\texttt{Conv2D}_{3 \times 3}(\texttt{MLP}(\mathbf{F}_{\text{in}})))) + \mathbf{F}_{\text{in}}, 
\end{equation}
where $\mathbf{F}_{\text{in}}$ are the features from self-attention, and $\texttt{GELU}$ denotes Gaussian Error Linear Unit activation. Our positional encoding scheme differs from the fixed positional encoding utilized in previous transformer networks like ViT \cite{vit} which allows our \ChangeFormer{} to take test images that are different in resolution from the ones used during training.

\subsubsection{Downsampling Block} 
Given an input patch $\mathbf{F}_i$ from the $i$-th transformer layer of resolution $\frac{H}{2^{i+1}} \times \frac{W}{2^{i+1}} \times C_i$, downsampling layer shrink it to obtain $\mathbf{F}_{i+1}$ of resolution $\frac{H}{2^{i+2}} \times \frac{W}{2^{i+2}} \times C_{i+1}$ which will be the input to the $(i+1)$-th Transformer layer. To achieve this, we utilize a $3 \times 3$ $\tt Conv2D$ layer with  kernel size $K = 7$, stride $S = 4$, and padding $P = 3$ for the initial downsampling, and  $K = 3$, $S = 2$, and $P = 1$ for the rest.

\subsubsection{Difference Module} 
We utilize four Difference Modules to compute the difference of multi-level features of pre-change and post-change images from the hierarchical transformer encoder as shown in Fig. \ref{fig:ChangeFormer}. More precisely, our Difference Module consists of \texttt{Conv2D, ReLU, BatchNorm2d (BN)} as follows:
\begin{equation}
    \mathbf{F}^i_{\text{diff}} = \texttt{BN}(\texttt{ReLU}(\texttt{Conv2D}_{3 \times 3}(\texttt{Cat}(\mathbf{F}^i_{\text{pre}},\mathbf{F}^i_{\text{post}})))),
\end{equation}
where $\mathbf{F}^i_{\text{pre}}$ and $\mathbf{F}^i_{\text{post}}$ denote the feature maps of pre-change and post-change images from the  $i$-th hierarchical layer, and $\texttt{Cat}$ denotes the tensor concatenation. Instead of computing the absolute difference of $\mathbf{F}^i_{\text{pre}}$ and $\mathbf{F}^i_{\text{post}}$ as in \cite{transformer_cd}, the proposed difference module learn the optimal distance metric at each scale during training - resulting in better CD performance.

\subsection{MLP Decoder}
We utilize a simple decoder with MLP layers that aggregates the multi-level feature difference maps to predict the change map. The proposed MLP decoder consists of three main steps.

\subsubsection{MLP \& Upsampling} 
We first process each multi-scale feature difference map through an MLP layer to  unify the channel dimension and then upsample each one to the size of $H/4 \times W/4$ as follows:
\begin{equation}
    \Tilde{\mathbf{F}}^i_{\text{diff}} = \texttt{Linear}(C_i, C_{\text{ebd}})(\mathbf{F}^i_{\text{diff}}) \forall i,
\end{equation}
\begin{equation}
    \hat{\mathbf{F}}^i_{\text{diff}} = \texttt{Upsample}((H/4, W/4), ``\text{bilinear}")(\Tilde{\mathbf{F}}^i_{\text{diff}}),
\end{equation}
where $C_{\text{ebd}}$ denotes the embedding dimension.

\subsubsection{Concatenation \& Fusion} 
The upsampled feature difference maps are then concatenated and fused through an MLP layer as follows:
\begin{equation}
    \mathbf{F} = \texttt{Linear}(4C_{\text{ebd}}, C_{\text{ebd}})(\texttt{Cat}(\hat{\mathbf{F}}^1_{\text{diff}},\hat{\mathbf{F}}^2_{\text{diff}},\hat{\mathbf{F}}^3_{\text{diff}},\hat{\mathbf{F}}^4_{\text{diff}})).
\end{equation}

\subsubsection{Upsampling \& Classification.} We upsample the fused feature map $\mathbf{F}$ to the size of $H\times W$ by utilizing a 2D transposed convolution layer with $S=4$ and $K=3$. Finally, the upsampled fused feature map is processed through another MLP layer to predict the
change mask $\mathbf{CM}$ with a resolution of $H \times W \times N_{\text{cls}}$, where $N_{\text{cls}}$ (=2) is the number of classes i.e., \textit{change} and \textit{no-change}. This process can be formulated as follows:
\begin{equation}
    \hat{\mathbf{F}} = \texttt{ConvTranspose2D}(S=4, K=3)(\mathbf{F}),
\end{equation}
\begin{equation}
    \mathbf{CM} = \texttt{Linear}(C_{\text{ebd}}, N_{\text{cls}})(\hat{\mathbf{F}}).
\end{equation}

\section{Experimental Setup}
\subsection{Datasets} 
We use two publically available CD datasets for our experiments, namely LEVIR-CD \cite{LEVIR} and DSIFN-CD \cite{DSIFN}. The LEVIR-CD is a building CD dataset that contains  RS image pairs of resolution $1024 \times 1024$. From these images, we crop non-overlapping patches of size $256 \times 256$ and randomly split them into three parts to make train/val/test sets of samples 7120/1024/2048. The DSIFN dataset is an general CD dataset that contains the changes in different land-cover objects. For experiments, we create non-overlapping patches of size $256 \times 256$ from the $512 \times 512$ images while utilizing the authors' default train/val/test sets. This results in 14400/1360/192 samples for training/val/test, respectively, for the DSIFN dataset. 

\subsection{Implementation Details} 
We implemented our model in PyTorch and trained using an NVIDIA Quadro RTX 8000 GPU. We randomly initialize the network. During training, we applied data augmentation through random flip, random re-scale (0.8-1.2), random crop, Gaussian blur, and random color jittering. We trained the models using the Cross-Entropy (CE) Loss and \texttt{AdamW} optimizer with weight decay equal to 0.01 and beta values equal to (0.9, 0.999). The learning rate is initially set to 0.0001 and linearly decays to 0 until trained for 200 epochs. We use a batch size of 16 to train the model. 

\subsection{Performance Metrics} To compare the performance of our model with SOTA methods, we report F1 and Intersection over Union (IoU) scores with regard to the \textit{change-class} as the primary quantitative indices. Additionally,  we report precision and recall of the change category and overall accuracy (OA).

\begin{table*}[htb]
    \caption{The average quantitative results of different CD methods on LEVIR-CD \cite{LEVIR} and DSIFN-CD~\cite{DSIFN}.*}
    \centering
    \begin{tabular}{lccccccccccc}
        \toprule
        \multirow{2}{*}{\centering Method} & \multicolumn{5}{c}{LEVIR-CD~\cite{LEVIR}} & & \multicolumn{5}{c}{DSIFN-CD~\cite{DSIFN}}\\
        \cmidrule{2-6} \cmidrule{8-12}
                            &       Precision & 
                                    Recall & 
                                    F1 & 
                                    IoU & 
                                    OA && 
                                    Precision & 
                                    Recall & 
                                    F1 & 
                                    IoU & 
                                    OA\\
        \toprule
        FC-EF~\cite{CD_FC} &        86.91 & 
                                    80.17 & 
                                    83.40 & 
                                    71.53 & 
                                    98.39 && 
                                    \textcolor{blue}{\bf 72.61} & 
                                    52.73 & 
                                    61.09 & 
                                    43.98 & 
                                    \bf 88.59\\
        FC-Siam-Di~\cite{CD_FC}     & 89.53 & 
                                    83.31 & 
                                    86.31 & 
                                    75.92 & 
                                    98.67 && 
                                    59.67 & 
                                    65.71 & 
                                    62.54 & 
                                    45.50 & 
                                    86.63\\
        FC-Siam-Conc~\cite{CD_FC}   & \bf 91.99 & 
                                    76.77 & 
                                    83.69 & 
                                    71.96 & 
                                    98.49 && 
                                    66.45 & 
                                    54.21 & 
                                    59.71 & 
                                    42.56 & 
                                    87.57\\
        DTCDSCN~\cite{CD_DTCDSCN}   & 88.53 & 
                                    86.83 & 
                                    87.67 & 
                                    78.05 & 
                                    98.77 && 
                                    53.87 & 
                                    \textcolor{blue}{\bf 77.99} & 
                                    63.72 & 
                                    46.76 & 
                                    84.91\\
        STANet~\cite{CD_STANet}     & 83.81 & 
                                    \textcolor{red}{\bf 91.00} & 
                                    87.26 & 
                                    77.40 & 
                                    98.66 && 
                                    67.71 & 
                                    61.68 & 
                                    64.56 & 
                                    47.66 & 
                                    88.49\\
        IFNet~\cite{CD_IFNet} &     \textcolor{red}{\bf 94.02} & 
                                    82.93 & 
                                    88.13 & 
                                    78.77 & 
                                    {\bf 98.87} && 
                                    67.86 & 
                                    53.94 & 
                                    60.10 & 
                                    42.96 & 
                                    87.83 \\
        SNUNet~\cite{CD_SNUNet} &   89.18 & 
                                    87.17 & 
                                    {\bf 88.16} & 
                                    {\bf 78.83} & 
                                    98.82 && 
                                    60.60 & 
                                    \textcolor{black}{\bf 72.89} & 
                                    \bf 66.18 & 
                                    \bf 49.45 & 
                                    87.34\\
        BIT~\cite{transformer_cd} & 89.24 & 
                                    \textcolor{blue}{\bf 89.37} & 
                                    \textcolor{blue}{\bf 89.31} & 
                                    \textcolor{blue}{\bf 80.68} & 
                                    \textcolor{blue}{\bf 98.92} && 
                                    \bf 68.36 & 
                                    70.18 & 
                                    \textcolor{blue}{\bf 69.26} & 
                                    \textcolor{blue}{\bf52.97} & 
                                    \textcolor{blue}{\bf89.41}\\
        \ChangeFormer{} (ours) &    \textcolor{blue}{\bf 92.05} & 
                                    {\bf 88.80} & 
                                    \textcolor{red}{\bf 90.40} & 
                                    \textcolor{red}{\bf 82.48} & 
                                    \textcolor{red}{\bf 99.04} && 
                                    \textcolor{red}{\bf 88.48} & 
                                    \textcolor{red}{\bf 84.94} & 
                                    \textcolor{red}{\bf 86.67} & 
                                    \textcolor{red}{\bf 76.48} & 
                                    \textcolor{red}{\bf 95.56}\\
        \bottomrule
        \multicolumn{12}{l}{{*All values are reported in percentage (\%). Color convention: \textcolor{red}{\bf best}, \textcolor{blue}{\bf 2nd-best}, and \textcolor{black}{\bf 3rd-best}.%
		}}
    \end{tabular}
    \label{tab:results}
\end{table*}
\begin{figure*}[tbh]
    \centering
    \includegraphics[width=\linewidth]{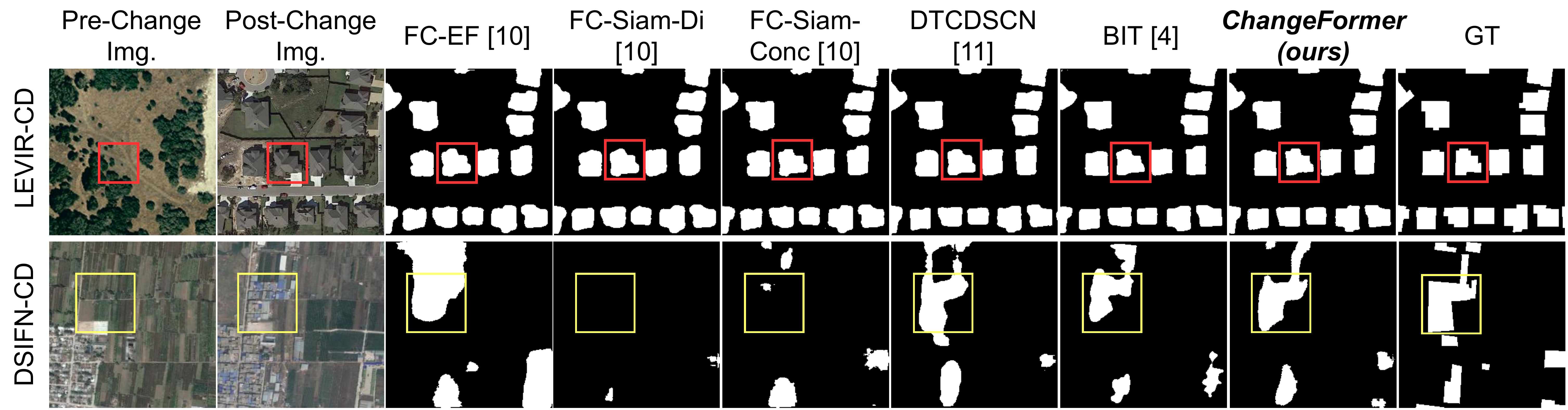}
    \caption{Qualitative results of different CD methods on LEVIR-CD~\cite{LEVIR} and DSIFN-CD~\cite{DSIFN}.}
    \label{fig:LEVIR}
\end{figure*}

\section{Results and Discussion}
\label{sec:results}
In this section, we compare the CD performance of our \textit{ChangeFormer} with existing SOTA methods:
\begin{itemize}[noitemsep]
    \itemsep0em
    \item \textbf{FC-EF}~\cite{CD_FC}: concatenates bi-temporal images and processes them through a ConvNet to detect changes. 
    \item \textbf{FC-Siam-Di}~\cite{CD_FC}: is a feature-difference method, which extracts multi-level features of bi-temporal images from a Siamese ConvNet, and  their difference is used to detect changes. 
    \item \textbf{FC-Siam-Conc}~\cite{CD_FC}: is a feature-concatenation method, which extracts multi-level features of bi-temporal images from a Siamese ConvNet, and feature concatenation is used to detect changes. 
    \item \textbf{DTCDSCN}~\cite{CD_DTCDSCN}: is an attention-based method, which utilizes a dual attention module (DAM) to exploit the inter-dependencies between channels and spatial positions of ConvNet features to detect changes. 
    \item \textbf{STANet}~\cite{CD_STANet}: is an another Siamese-based spatial-temporal attention network for CD.
    \item \textbf{IFNet}~\cite{CD_IFNet}: is a multi-scale feature concatenation method, which fuses multi-level deep features of bi-temporal images with image difference features by means of attention modules for change map reconstruction.
    \item \textbf{SNUNet}~\cite{CD_SNUNet}: is a multi-level feature concatenation method, in which a densely connected (NestedUNet) Siamese network is used for change detection.
    \item \textbf{BIT}~\cite{transformer_cd}: is a transformer-based method, which uses a transformer encoder-decoder network to enhance the context-information of ConvNet features via semantic tokens followed by feature differencing to obtain the change map. 
\end{itemize}

\par Table \ref{tab:results} presents the results of different CD methods on the test-sets of LEVIR-CD~\cite{LEVIR} and DSIFN-CD~\cite{DSIFN}. As can be seen from the table, the proposed \textit{ChangeFormer} network achieves better CD performance in terms of F1, IoU, and OA metrics. In particular, our \textit{ChangeFormer} improves previous SOTA in F1/IoU/OA by ~1.2/2.2/0.1\% and ~20.0/44.3/6.4\% for LEVIR-CD and DSIFN-CD, respectively. In addition, Fig. \ref{fig:LEVIR} compares the visual quality of different SOTA methods on test images from LEVIR-CD and DSIFN-CD. As highlighted in red, our ChangeFormer captures much finer details compared to the other SOTA methods. These quantitative and qualitative comparisons show the superiority of our proposed CD method over the existing SOTA methods.

\section{Conclusion}
\label{sec:con}
In this paper, we proposed a transformer-based Siamese network for CD. By utilizing a hierarchical transformer encoder in a Siamese architecture with a simple MLP decoder, our method outperforms several other recent CD methods that employ very large ConvNets like ResNet18 and U-Net as the backbone. We also show better performance in terms of IoU, F1 score, and overall accuracy than recent ConvNet-based (FC-EF, FC-Siam-DI, and FC-Siam-Conc), attention-based (DTCDSCN, STANet, and IFNet), and ConvNet+Transformer-based (BIT) methods. Hence, this study shows that it is unnecessary to depend on deep-ConvNets, and a hierarchical transformer in a Siamese network with a lightweight decoder can work very well for CD.

\section{Acknowledgment}
This work was supported by NSF CAREER award 2045489.

\let\oldthebibliography\thebibliography
\let\endoldthebibliography\endthebibliography
\renewenvironment{thebibliography}[1]{
  \begin{oldthebibliography}{#1}
    \setlength{\itemsep}{0em}
    \setlength{\parskip}{0em}
}
{
  \end{oldthebibliography}
}

\bibliography{refs}   

\begin{thebibliography}{10}

\bibitem{bandara2022revisiting}
Wele Gedara~Chaminda Bandara and Vishal~M Patel,
\newblock ``Revisiting consistency regularization for semi-supervised change
  detection in remote sensing images,''
\newblock {\em arXiv preprint arXiv:2204.08454}, 2022.

\bibitem{bandara2021spin}
Wele Gedara~Chaminda Bandara, Jeya Maria~Jose Valanarasu, and Vishal~M Patel,
\newblock ``Spin road mapper: Extracting roads from aerial images via spatial
  and interaction space graph reasoning for autonomous driving,''
\newblock {\em arXiv preprint arXiv:2109.07701}, 2021.

\bibitem{cd_attention}
Qian Shi, Mengxi Liu, Shengchen Li, Xiaoping Liu, Fei Wang, and Liangpei Zhang,
\newblock ``A deeply supervised attention metric-based network and an open
  aerial image dataset for remote sensing change detection,''
\newblock {\em IEEE Transactions on Geoscience and Remote Sensing}, 2021.

\bibitem{transformers_review}
Salman Khan, Muzammal Naseer, Munawar Hayat, Syed~Waqas Zamir, Fahad~Shahbaz
  Khan, and Mubarak Shah,
\newblock ``Transformers in vision: A survey,''
\newblock {\em arXiv preprint arXiv:2101.01169}, 2021.

\bibitem{segformer}
Enze Xie, Wenhai Wang, Zhiding Yu, Anima Anandkumar, Jose~M Alvarez, and Ping
  Luo,
\newblock ``Segformer: Simple and efficient design for semantic segmentation
  with transformers,''
\newblock {\em arXiv preprint arXiv:2105.15203}, 2021.

\bibitem{transformer_cd}
Hao Chen, Zipeng Qi, and Zhenwei Shi,
\newblock ``Remote sensing image change detection with transformers,''
\newblock {\em IEEE Transactions on Geoscience and Remote Sensing}, 2021.

\bibitem{original_attention}
Ashish Vaswani, Noam Shazeer, Niki Parmar, Jakob Uszkoreit, Llion Jones,
  Aidan~N Gomez, {\L}ukasz Kaiser, and Illia Polosukhin,
\newblock ``Attention is all you need,''
\newblock in {\em Advances in neural information processing systems}, 2017, pp.
  5998--6008.

\bibitem{eff_atten}
Wenhai Wang, Enze Xie, Xiang Li, Deng-Ping Fan, Kaitao Song, Ding Liang, Tong
  Lu, Ping Luo, and Ling Shao,
\newblock ``Pyramid vision transformer: A versatile backbone for dense
  prediction without convolutions,''
\newblock {\em arXiv preprint arXiv:2102.12122}, 2021.

\bibitem{vit}
Alexey Dosovitskiy, Lucas Beyer, Alexander Kolesnikov, Dirk Weissenborn,
  Xiaohua Zhai, Thomas Unterthiner, Mostafa Dehghani, Matthias Minderer, Georg
  Heigold, Sylvain Gelly, et~al.,
\newblock ``An image is worth 16x16 words: Transformers for image recognition
  at scale,''
\newblock {\em arXiv preprint arXiv:2010.11929}, 2020.

\bibitem{LEVIR}
Hao Chen and Zhenwei Shi,
\newblock ``A spatial-temporal attention-based method and a new dataset for
  remote sensing image change detection,''
\newblock {\em Remote Sensing}, vol. 12, no. 10, pp. 1662, 2020.

\bibitem{DSIFN}
Chenxiao Zhang, Peng Yue, Deodato Tapete, Liangcun Jiang, Boyi Shangguan,
  Li~Huang, and Guangchao Liu,
\newblock ``A deeply supervised image fusion network for change detection in
  high resolution bi-temporal remote sensing images,''
\newblock {\em ISPRS Journal of Photogrammetry and Remote Sensing}, vol. 166,
  pp. 183--200, 2020.

\bibitem{CD_FC}
Rodrigo~Caye Daudt, Bertr Le~Saux, and Alexandre Boulch,
\newblock ``Fully convolutional siamese networks for change detection,''
\newblock in {\em 2018 25th IEEE International Conference on Image Processing
  (ICIP)}. IEEE, 2018, pp. 4063--4067.

\bibitem{CD_DTCDSCN}
Yi~Liu, Chao Pang, Zongqian Zhan, Xiaomeng Zhang, and Xue Yang,
\newblock ``Building change detection for remote sensing images using a
  dual-task constrained deep siamese convolutional network model,''
\newblock {\em IEEE Geoscience and Remote Sensing Letters}, vol. 18, no. 5, pp.
  811--815, 2020.

\bibitem{CD_STANet}
Hao Chen and Zhenwei Shi,
\newblock ``A spatial-temporal attention-based method and a new dataset for
  remote sensing image change detection,''
\newblock {\em Remote Sensing}, vol. 12, no. 10, pp. 1662, 2020.

\bibitem{CD_IFNet}
Chenxiao Zhang, Peng Yue, Deodato Tapete, Liangcun Jiang, Boyi Shangguan,
  Li~Huang, and Guangchao Liu,
\newblock ``A deeply supervised image fusion network for change detection in
  high resolution bi-temporal remote sensing images,''
\newblock {\em ISPRS Journal of Photogrammetry and Remote Sensing}, vol. 166,
  pp. 183--200, 2020.

\bibitem{CD_SNUNet}
Sheng Fang, Kaiyu Li, Jinyuan Shao, and Zhe Li,
\newblock ``Snunet-cd: A densely connected siamese network for change detection
  of vhr images,''
\newblock {\em IEEE Geoscience and Remote Sensing Letters}, 2021.

\end{thebibliography}
\bibliographystyle{IEEEbib}


\end{document}